\documentclass[11pt]{article}

\usepackage[margin=1in]{geometry}
\usepackage{cite}
\usepackage{graphicx}
\usepackage{amsmath,amssymb,amsfonts}
\usepackage{array}
\usepackage{booktabs}
\usepackage{multirow}
\usepackage{textcomp}
\usepackage{url}
\usepackage{float}

\graphicspath{{figures/}{./}}
\begin{document}

\title{Multi-Subject Pretraining Enables Short-Calibration Personalization for Closed-Corpus Surface EMG Speech Decoding}

\author{%
Chenqian Le\textsuperscript{1},
Beatrice Fumagalli\textsuperscript{1,2},
Yasamin Esmaeili\textsuperscript{2,3},
Xupeng Chen\textsuperscript{1},\\
Tianyu He\textsuperscript{1},
Nikasadat Emami\textsuperscript{1},
Adeen Flinker\textsuperscript{2,3},
and Yao Wang\textsuperscript{1,3}\\[0.75em]
\parbox{0.9\textwidth}{\centering\small
\textsuperscript{1}Department of Electrical and Computer Engineering, New York University Tandon School of Engineering, New York, NY, USA\\
\textsuperscript{2}Department of Neurology, New York University Grossman School of Medicine, New York, NY, USA\\
\textsuperscript{3}Department of Biomedical Engineering, New York University Tandon School of Engineering, New York, NY, USA\\[0.5em]
Corresponding author: Yao Wang (yaowang@nyu.edu)
}
}
\date{}

\maketitle

\begin{abstract}
Surface electromyography (sEMG)-based silent speech interfaces are limited by cross-user variability and calibration burden. We study a limited-data setting in which each of 27 speech-typical participants contributed less than 0.5 h of data (21.3 min on average) across Aloud and Mimed speech. Within a closed 50-sentence corpus, we used leave-one-subject-out evaluation, initializing from a released single-subject checkpoint, pretraining on non-held-out participants, and fine-tuning on the target participant. This pipeline achieved 21.7\% character error rate (CER) and 31.9\% word error rate (WER), compared with 49.3\% CER without target-subject calibration and 68.0\% CER for direct checkpoint fine-tuning. Multi-subject pretraining from random initialization followed by fine-tuning reached 44.9\% CER and did not converge under the fixed schedule in 5 of 27 folds, indicating substantial optimization and accuracy benefits from checkpoint initialization. Macro-averaged CER declined from 74.4\% with one pretraining participant to 21.7\% with 26. Three minutes of target-subject calibration achieved 20.5\% CER and 31.7\% WER, with no statistically significant difference from the full approximately 13-min pool (21.7\% CER and 31.9\% WER). A subject-specific adapter provided no detectable benefit. Excluding the five evaluation sentences from all sEMG model-training data increased CER and WER to 78.6\% and 99.9\%. These results support short-calibration personalization in a standardized-montage, closed-corpus setting.
\end{abstract}


\section{Introduction}

Silent speech interfaces aim to decode intended speech from non-acoustic biosignals and may enable communication when audible speech is unavailable or impaired \cite{denby2010ssi,jorgensen2010semg_interfaces,schultz2017survey,silva2024speech_review}. Among candidate sensing modalities, surface electromyography (sEMG) is attractive because it is non-invasive, portable, and directly captures articulatory muscle activity. These properties make sEMG a promising substrate for assistive communication and user-adaptive rehabilitation technologies.

Despite this promise, robust speech decoding from sEMG remains difficult. Prior sEMG systems have demonstrated phoneme recognition, word recognition, and speech synthesis from facial muscle activity \cite{meltzner2011signal,janke2017emg,meltzner2017laryngectomy,gaddy2020digitalvoicing,gaddy-klein-2021-improved}. However, many sEMG systems still rely on subject-specific training using a substantial amount of data \cite{gaddy2020digitalvoicing,gaddy-klein-2021-improved,chen2023hdsemg}. Cross-speaker pretraining followed by adaptation to an unseen speaker has already been demonstrated for EMG-to-speech conversion \cite{scheck2024crossspeaker}, and recent work has explicitly evaluated cross-subject model calibration using different amounts of target-user data \cite{zeng2025crosssubject}. Other studies have pursued calibration-free cross-subject recognition through domain alignment \cite{zhang2023crosssubject,cui2024adversarial}. Our contribution is therefore not the pretrain-then-fine-tune paradigm itself. Rather, we study a complementary limited-data regime in which every participant contributes less than 0.5 h of data, and we systematically characterize two practical data-scaling factors within the same held-out-subject framework: the number of subjects available for multi-subject pretraining, and the amount of labeled calibration data available from a new user. A practical interface should reuse information learned from other users while requiring only a short calibration session.

In this study, we focus on cross-subject transfer for sentence-level sEMG speech decoding within a closed 50-sentence corpus. Our study cohort consists of 27 speech-typical participants who each contributed an average of 21.3 min of sEMG data across the Aloud and Mimed conditions. We ask six questions through  leave-one-subject-out analyses. First, can a multi-subject model pretrained on other subjects be personalized to a held-out subject with short calibration? Second, does this personalized model outperform transfer without target-subject calibration (zero-shot transfer) and direct fine-tuning of the released single-subject model? Third, how does performance change as the number of pretraining subjects increases? Fourth, does a subject-specific multilayer perceptron (MLP) adapter improve personalization? Fifth, how much target-subject calibration data are needed? Sixth, does the adapted model generalize to sentences not seen during training?
\begin{figure*}[!t]
\centering
\includegraphics[width=\textwidth]{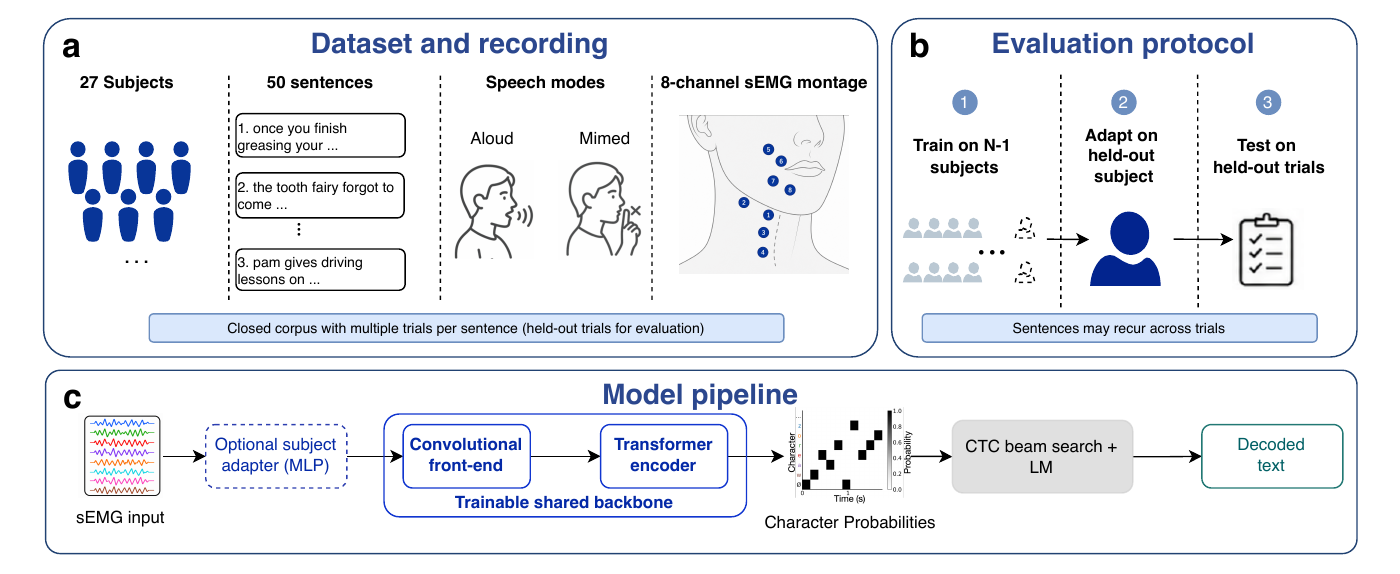}
\caption{Overview of the cross-subject sEMG speech decoding framework. (a) Dataset and recording: 27 subjects, a 50-sentence corpus, Aloud and Mimed speech modes, and an eight-channel sEMG montage over the lower face and neck. (b) Closed-corpus LOSO personalization protocol. A single fixed 45/5 split of the corpus is shared by all 27 subjects: the same five evaluation sentences are held out for every subject. Target-subject calibration uses trials from the remaining 45 sentences, while one Aloud and one Mimed trial per evaluation sentence is reserved for testing when available, giving a nominal 10-trial test set per subject. Multi-subject pretraining uses training-partition trials from the 26 non-held-out subjects, which may include the non-reserved repetitions of the evaluation sentences. The primary analysis and the unseen-sentence test both consider all 27 subjects. Primary decoding metrics pool Aloud and Mimed trials. (c) Model pipeline: the model maps surface EMG to frame-level character probabilities, and CTC beam search with a language model converts them into text. The subject-specific MLP adapter is optional and is evaluated in the ablation analysis.}
\label{fig:main_pipeline}
\end{figure*}
The main contributions are:
\begin{enumerate}
    \item We show that multi-subject pretraining followed by target-subject fine-tuning substantially outperforms zero-shot transfer and direct target-subject fine-tuning of the released checkpoint under the same training schedule.
    \item Under the fixed-epoch schedule, 3 min of target-subject calibration yielded error rates that were not significantly different from those obtained with the full approximately 13-min calibration pool.
    \item Held-out decoding accuracy improved substantially as the number of pretraining subjects increased, from 74.4\% CER with one subject to 21.7\% with 26, showing that a larger multi-subject pretraining cohort supports more effective personalization.
    \item A subject-specific MLP adapter provided no detectable benefit, while decoding degraded sharply when the evaluation sentences were absent from all sEMG model-training data.
\end{enumerate}

\section{Related Work}

\subsection{Silent Speech Interfaces}
Silent speech interfaces seek to recover linguistic content from non-acoustic signals, including ultrasound, electromagnetic articulography, EEG, ECoG, and EMG \cite{denby2010ssi,jorgensen2010semg_interfaces,schultz2017survey}. In rehabilitation engineering, these systems are motivated by communication support for users whose audible speech is impaired or unavailable. In neural engineering, they provide a testbed for decoding articulatory and linguistic representations from biosignals. Within this broader area, surface EMG is attractive because it is non-invasive and wearable, but it is also highly sensitive to electrode placement, muscle recruitment, speaking mode, and subject physiology.

\subsection{sEMG-Based Speech Decoding}
Surface EMG has been used for silent speech recognition and EMG-to-speech synthesis because it records neuromuscular activity related to speech production \cite{jou2006continuous,schultz2010coarticulation,meltzner2011signal,wand2014emguka,wand2014tackling,janke2017emg,meltzner2017laryngectomy,li2023semg_review}. Early systems established continuous EMG speech recognition with phonetic modeling and handcrafted features \cite{jou2006continuous,schultz2010coarticulation}, while later work examined speaking-mode mismatch between audible and silent speech \cite{wand2014emguka,wand2014tackling}. Recent neural approaches use learned temporal features and sequence objectives to decode or synthesize from EMG, including DNN frontends for continuous EMG speech recognition \cite{wand2016dnnfrontend}, high-density sEMG classifiers \cite{chen2023hdsemg,cui2024adversarial}, and open-vocabulary silent speech voicing systems \cite{gaddy2020digitalvoicing,gaddy-klein-2021-improved}. More broadly, large-scale non-speech sEMG decoding has recently shown that cross-user generalization and personalization can be improved with large participant pools and modern sequence models \cite{kaifosh2025generic}. However, subject variability, electrode placement differences, and session-level drift remain central barriers. Cross-speaker EMG-to-speech models have previously been adapted to leave-out speakers after training on other speakers \cite{scheck2024crossspeaker}, and Zeng et al. explicitly evaluated multiple target-subject calibration scales in an 11-subject cross-subject sEMG-to-speech system \cite{zeng2025crosssubject}. Cross-subject silent-speech recognition has also been studied with domain-adversarial approaches, including large-cohort sentence-classification experiments \cite{zhang2023crosssubject,cui2024adversarial}. These studies establish the feasibility of cross-subject transfer and adaptation, but they address different parts of the data-scaling problem and mostly target speech synthesis or fixed-sentence classification. To our knowledge, prior work has not systematically evaluated both pretraining-cohort size and target-subject calibration duration within the same held-out-subject character-sequence decoding framework, in which the decoder emits a character sequence under a connectionist temporal classification objective and is scored by character error rate.

\subsection{Cross-Subject Transfer and Personalization}
Cross-subject learning can reduce the burden of collecting large amounts of user-specific data. The difficulty is that EMG amplitude, spatial distribution, timing, and noise structure vary across users. Prior work has addressed related adaptation problems at the session or speaker level using supervised and unsupervised adaptation, domain-adversarial training, meta-learning, and cross-speaker EMG-to-speech conversion \cite{maierhein2005session,wand2014unsupervised,ganin2015domain_adaptation,wand2018domain,finn2017maml,prorokovic2019meta,scheck2024crossspeaker}. Recent cross-subject silent-speech work has also used adversarial alignment to reduce inter-subject domain offsets \cite{cui2024adversarial}. Personalization is therefore important, but the practical question is how much adaptation data is needed, how much multi-subject pretraining helps, and which model components should be updated.

\subsection{Study Scope}
Evidence remains limited on how a shared sEMG decoder changes as data from more subjects are added and how much calibration a new user needs. We examine both questions in a 27-subject closed-corpus LOSO analysis. The unseen-sentence test is included only to define where those personalization results apply.

\section{Methods}
 
\subsection{Participants and Data Acquisition}

Surface EMG data were collected from 27 speech-typical participants during sentence production tasks. The experimental procedures were approved by New York University's Institutional Review Board under the protocol ``Auditory Sequences: Psychophysics and Neurophysiology'' (IRB-FY2020-4157). The recording corpus was based on 50 distinct sentences drawn from the TIMIT corpus \cite{garofolo1993timit}. Each sentence was nominally repeated three times per speech mode (Aloud, Mimed and Subvocal). Surface EMG was recorded from lower-face and neck muscle regions using the same eight-channel electrode locations as in \cite{meltzner2011signal}. All 27 participants of the present cohort were recorded with this same nominal eight-channel electrode-placement protocol, so the cross-subject results below are obtained under a standardized montage rather than across heterogeneous sensor layouts. The released single-subject checkpoint used for initialization was recorded with a different eight-channel montage \cite{gaddy2020digitalvoicing}. The channels are numbered in Fig.~\ref{fig:main_pipeline}(a), and their anatomical placements are listed in Supplementary Table~S2.


The analysis used data from all 27 participants. Subvocal trials were not used, leaving up to 300 Aloud and Mimed trials per subject before unusable recordings were excluded.


\subsection{Training and Testing Split}
\label{sec:eval_split}

A single fixed set of five evaluation sentences was selected from the 50-sentence corpus and used for all 27 participants, giving a fixed 45/5 split into calibration and evaluation sentences. For each participant, one Aloud and one Mimed trial was reserved for each evaluation sentence when available. This yielded a nominal test set of 10 trials per participant and 269 held-out trials in total, because one Aloud trial was unavailable for participant P7. The evaluation sentences and reserved test trials were fixed across all methods. Details of how the five evaluation sentences were selected are given in the supplementary material.

Each LOSO fold comprised multi-subject pretraining and, where applicable, target-subject calibration. During pretraining, the model used the training-partition trials from the 26 non-held-out subjects. Specifically, for each non-held-out subject, the trials reserved for that subject's own evaluation set were excluded. All other usable Aloud and Mimed trials were retained, including the remaining repetitions of the five evaluation sentences. This provided on average 282 usable trials per non-held-out subject.

For target-subject calibration, all repetitions of the held-out subject's five evaluation sentences were excluded. Full calibration used the usable Aloud and Mimed trials from the remaining 45 sentences, corresponding to approximately 263 trials and 13 min of fine-tuning data on average. The non-reserved repetitions of the five evaluation sentences were not used for target-subject calibration or testing. Thus, in the primary closed-corpus analysis, the evaluation sentences were absent from target-subject calibration but could appear in pretraining data from other subjects.

The direct Checkpoint + fine-tune experiments used a broader target-subject pool: they excluded the reserved test trials but retained the remaining repetitions of the five evaluation sentences, yielding approximately 282 trials and 14 min on average. Thus, these control experiments shared the test trials and optimization hyperparameters of the primary comparisons but were not strictly matched in calibration data.

We also ran an unseen-sentence experiment to test whether the model generalizes to sentences it has never seen. It used the same reserved test trials per held-out subject but additionally excluded all trials of the five evaluation sentences from multi-subject pretraining. The evaluation sentences for a held-out subject were therefore absent from both pretraining and target-subject calibration.

\subsection{Decoding Preprocessing}
\label{sec:decoding_preprocessing}

For decoding experiments, raw sEMG signals were preprocessed using the decoding-oriented pipeline adopted in the \texttt{silent\_speech} framework \cite{gaddy2020digitalvoicing,gaddy-klein-2021-improved}. Signals were resampled from 2222.22 Hz to 1000 Hz via polyphase resampling. Each channel was filtered with notch filters at 60 Hz and harmonics up to 420 Hz, followed by a third-order high-pass Butterworth filter with a 2 Hz cutoff. The denoised 1000 Hz signals were then downsampled to 689 Hz using linear interpolation. The resulting signals were used as model input.

\subsection{Decoding Model and Training Conditions}
\label{sec:models_training}

The decoding architecture follows the convolution--transformer model of Gaddy and Klein for mapping silent sEMG to characters \cite{gaddy2020digitalvoicing,gaddy-klein-2021-improved}. It predicts character sequences with a connectionist temporal classification (CTC) objective. We start with their released weights, trained on approximately 18.7 hours of EMG data from one subject (15.1 hours of vocalized speech and 3.6 hours of silent speech); we refer to this as checkpoint initialization.

This initial model is first refined using data from non-held-out subjects and then fine-tuned on the held-out subject's calibration data. We call the first stage multi-subject pretraining. To account for signal variation across subjects, we also examined a residual MLP adapter placed before the shared convolution--transformer backbone, as shown in Fig.~\ref{fig:main_pipeline}(c). For subject $s$, the adapter computes $y=x+\alpha_s f(x; \theta_s)$, where $f (x;{\theta_s})$ comprises LayerNorm over the eight input channels $x$ followed by an $8\rightarrow128\rightarrow8$ two-layer mapping. Each adapter contains 2,201 trainable parameters, including the learned residual scale $\alpha_s$, which is initialized to $10^{-3}$. During multi-subject pretraining, each non-held-out subject has separate adapter parameters $(\theta_s,\alpha_s)$ while the backbone is shared. For the held-out subject, a new adapter is instantiated; no adapter parameters from any pretraining subject are reused. In calibrated conditions, this new adapter and the shared backbone are fine-tuned jointly on the held-out subject's calibration trials.


We report results from the following model variants on held-out subject test data:
\begin{enumerate}
    \item \textbf{Pretrain + fine-tune (MLP):} The convolution-transformer backbone is augmented with the MLP adapter. The adapter is subject-specific, but the  backbone is shared among all subjects, initialized by the released checkpoint weights. The adapters and the backbone are first trained  on non-held-out subjects, and then fine-tuned using held-out-subject calibration data.
    \item \textbf{Pretrain + fine-tune (no MLP):} The original backbone checkpoint is first refined using non-held-out subject data without subject-specific MLPs, and then  fine-tuned on held-out-subject calibration data.
    \item \textbf{Pretrain zero-shot (no MLP):} The original backbone checkpoint is first refined using non-held-out subject data, and evaluated directly on held-out-subject test trials.
    \item \textbf{Pretrain (from random init) zero-shot (no MLP):} The backbone model is trained from random initialization on non-held-out subjects and evaluated directly on the held-out subject.
    \item \textbf{Pretrain (from random init) + fine-tune (MLP):} Identical to Pretrain + fine-tune (MLP), except that the shared backbone is randomly initialized rather than initialized from the released single-subject checkpoint. The model is first pretrained on the non-held-out subjects and then fine-tuned on the held-out subject using the same data split and optimization schedule.
    \item \textbf{Checkpoint + fine-tune (MLP and no MLP):} The backbone checkpoint is directly fine-tuned on the broader target-subject pool defined in Section~\ref{sec:eval_split}, with or without the MLP adapter (the two Checkpoint + fine-tune rows of Table~\ref{tab:full_cohort_benchmark}).
\end{enumerate}
The Checkpoint + fine-tune experiment  tests whether performance gains are explained merely by checkpoint initialization plus target-subject fine-tuning, without  multi-subject pretraining. The Pretrain (from random init) + fine-tune experiment is the converse control: it removes the released checkpoint while keeping multi-subject pretraining and target-subject fine-tuning, and therefore isolates what the released initialization contributes to the final personalized model. The no-MLP ablation tests whether the subject-specific MLP  provides benefit beyond the target-subject fine-tuning.

Architecture, training, and decoder hyperparameters were fixed before test evaluation. All conditions were optimized with AdamW at a base learning rate of $3\times10^{-4}$ and no weight decay, using a linear warm-up over the first 100 optimizer steps followed by a step schedule that halved the learning rate after epochs 125, 150, and 175. The batch size was 128 utterances on a single GPU, so the batch size is also the effective batch size, and no gradient clipping was applied. Both LOSO stages, multi-subject pretraining and target-subject fine-tuning, used these same optimization settings. Main experiments were trained for 400 epochs, and the unseen-sentence models were trained for 800 epochs. In every condition, we report the model at the final epoch.
We did not tune decoder or training hyperparameters separately for individual held-out subjects based on test-set WER or CER.

Decoding used connectionist temporal classification (CTC) \cite{graves2006ctc,sun2020brain2char} beam search with language-model refinement via \texttt{pyctcdecode} and an external KenLM language model \cite{heafield2011kenlm}. The language model was not refined on the 50-sentence inventory used in the present experiments. Beam decoding used \texttt{pyctcdecode} with beam width 100 and fixed decoder weights $\alpha_{\mathrm{LM}}=0.5$ and $\beta_{\mathrm{LM}}=0.5$. Main results report macro-averaged character error rate (CER) and word error rate (WER) across held-out subjects' test trials (both Aloud and Mimed).

\subsection{Calibration, Unseen Sentences, and Pretraining Pool Size}

To quantify calibration efficiency, we evaluated target-subject calibration budgets of 1, 3, 5, and 10 min for the Pretrain + fine-tune (MLP) model across all 27 LOSO folds. Fine-tuning with the complete 45-sentence calibration pool, approximately 263 usable trials and 13 min on average, is reported as a reference. Candidate trials from this pool were shuffled with a fixed random seed and included until the requested duration was reached. The selection was not stratified by sentence or speech mode. Because the available data varied across subjects, this is a time-budget comparison rather than a fixed-trial-count comparison.

To evaluate the potential benefit of including the MLP adapter, we repeated full target-subject fine-tuning with and without this module across all 27 subjects. To test generalization to new content, we additionally removed each held-out subject's five evaluation sentences from multi-subject pretraining; these sentences were already excluded from target-subject calibration. To measure the effect of the pretraining pool size, we used $n \in \{1,3,5,15,20,26\}$ pretraining subjects and evaluated all 27 held-out subjects at every pool size. For each fold and pool size, the pretraining pool was randomly selected from the remaining 26 non-held-out subjects using a fixed seed. The $n=26$ point is the Pretrain + fine-tune (MLP) result in Table~\ref{tab:full_cohort_benchmark}. This analysis used the MLP model at every pool size, so it does not test whether the adapter's effect changes with pool size.

\subsection{Statistical Analysis}
\label{sec:stats}

CER and WER are macro-averaged across held-out subjects unless stated otherwise. We used subject-level paired, two-sided Wilcoxon signed-rank tests \cite{wilcoxon1945} for the 27-subject protocol controls, the adapter comparison, and each calibration budget versus full fine-tuning. CER and WER were tested separately. The calibration-budget comparisons are descriptive and are reported as uncorrected $p$ values, without multiple-comparison correction across budgets. Confidence intervals on paired differences are 95\% percentile intervals from a 10{,}000-sample bootstrap that resamples held-out subjects and recomputes the mean paired difference; correlation intervals instead use the Fisher-$z$ transform. Error-rate differences are absolute percentage-point differences unless identified as relative changes.

For the random-initialization control, a fold was classified as non-converged when the pretraining stage did not improve on its initial error over the full fixed epoch budget and the model remained in a degenerate connectionist temporal classification solution, so that the held-out predictions after target-subject fine-tuning were empty and yielded 100\% CER and WER. These values are measured outcomes rather than imputed penalties. The primary paired comparison retains the observed outcomes from all 27 folds; we additionally report a sensitivity analysis restricted to the folds that converged.

The calibration and adapter figures show subject-level distributions: boxes give the interquartile range and median, and dots show individual subjects. Fig.~\ref{fig:full_cohort} reports macro means with $\pm$1 Std error bars and overlays individual subjects. Group comparisons that are not subject-paired, such as the encoding-score comparison in Section~\ref{sec:results_encdec_corr}, used two-sided Mann--Whitney tests; correlation confidence intervals use the Fisher-$z$ transform.

\section{Results}

\subsection{Closed-Corpus LOSO Analysis Across 27 Subjects}
\label{sec:results_full_cohort}

\begin{figure*}[!t]
\centering
\includegraphics[width=\textwidth]{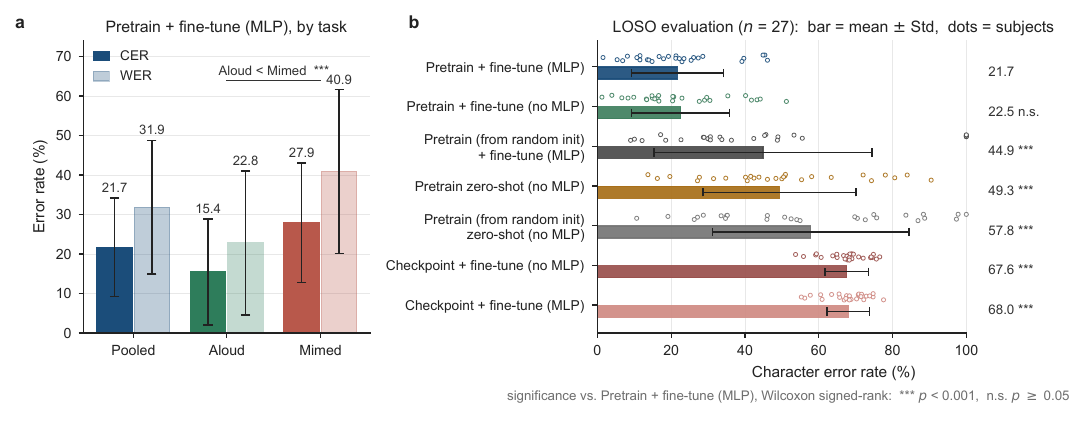}
\caption{LOSO evaluation of 27 held-out subjects. (a) Pooled, Aloud, and Mimed CER and WER for Pretrain + fine-tune (MLP); bars show macro means $\pm$1 subject-level Std. (b) Pooled CER for the seven conditions in Table~\ref{tab:full_cohort_benchmark}; bars show macro means $\pm$1 Std, and dots show subjects. In the Pretrain (from random init) + fine-tune (MLP) row, the five subjects at 100\% CER are the folds in which multi-subject pretraining did not converge. Stars denote paired two-sided Wilcoxon comparisons against Pretrain + fine-tune (MLP) ($^{***}p<0.001$; n.s.\ $p\geq0.05$). The bracket denotes the Aloud-versus-Mimed CER comparison ($^{***}p=3.0\times10^{-5}$).}
\label{fig:full_cohort}
\end{figure*}

Table~\ref{tab:full_cohort_benchmark} and Figure~\ref{fig:full_cohort} summarize the results from all model variants. All calibration data were used for target-subject fine-tuning.
Multi-subject pretraining followed by supervised target-subject fine-tuning achieved the lowest errors, with the MLP variant reaching 21.7\% CER / 31.9\% WER. Both pretrained zero-shot variants remained substantially worse, as did direct fine-tuning of the initial checkpoint without multi-subject pretraining. The Checkpoint + fine-tune condition is a protocol control showing that, under the same optimization schedule, checkpoint initialization followed by target-subject-only fine-tuning was insufficient to reproduce the benefit of multi-subject pretraining. It used the same optimization hyperparameters but the broader target-subject pool described in Section~\ref{sec:eval_split}, and it is not intended as a comparison against a separately optimized single-subject system.

\begin{table}[!ht]
\renewcommand{\arraystretch}{1.15}
\caption{LOSO evaluation ($n=27$). CER and WER are macro means over the 27 held-out subjects, pooling Aloud and Mimed trials; Std is the subject-level sample standard deviation.}
\label{tab:full_cohort_benchmark}
\centering
\footnotesize
\resizebox{\columnwidth}{!}{%
\begin{tabular}{lccccc}
\toprule
Condition & $n$ & CER (\%) & CER Std & WER (\%) & WER Std \\
\midrule
Pretrain + fine-tune (no MLP) & 27 & 22.5 & 13.3 & 33.3 & 18.3 \\
Pretrain + fine-tune (MLP) & 27 & \textbf{21.7} & 12.5 & \textbf{31.9} & 17.0 \\
Pretrain (from random init) + fine-tune (MLP)$^{\dagger}$ & 27 & 44.9 & 29.5 & 58.8 & 26.0 \\
Pretrain zero-shot (no MLP)& 27 & 49.3 & 20.8 & 63.1 & 21.1 \\
Pretrain (from random init) zero-shot (no MLP)& 27 & 57.8 & 26.7 & 70.7 & 24.5 \\
Checkpoint + fine-tune (no MLP) & 27 & 67.6 & 5.9 & 96.7 & 5.0 \\
Checkpoint + fine-tune (MLP) & 27 & 68.0 & 5.8 & 97.2 & 4.3 \\
\bottomrule
\end{tabular}}
\begin{flushleft}
\footnotesize
$^{\dagger}$This condition is identical to Pretrain + fine-tune (MLP) except that the shared backbone starts from random weights instead of the released checkpoint. Multi-subject pretraining did not converge in 5 of the 27 folds; those models produced 100\% CER and WER, and all 27 observed outcomes are included here. A sensitivity analysis restricted to the converged folds is given in Section~\ref{sec:results_full_cohort}.
\end{flushleft}
\end{table}

Supplementary Table~S1 and Fig.~\ref{fig:full_cohort} provide the paired comparisons supporting two findings. First, pretraining using data from multiple subjects brings a substantial gain when only limited target-subject calibration data are available; direct checkpoint fine-tuning performs poorly despite using the broader target-subject pool described in Section~\ref{sec:eval_split}. Second, target-subject fine-tuning is necessary because zero-shot transfer from a multi-subject model remains far from the fine-tuned models. Mimed trials remained harder than Aloud trials (27.9\% vs.\ 15.4\% CER; paired Wilcoxon $p=3.0\times10^{-5}$).

To isolate the contribution of the released single-subject checkpoint, we repeated the full pipeline with the shared backbone initialized randomly instead, keeping the architecture, schedule, and data splits identical. Multi-subject pretraining from random initialization was unstable at this data scale: in 5 of the 27 folds it did not converge, and the resulting models produced 100\% CER and WER on the held-out subject. Over all 27 folds, this condition reached 44.9\% CER and 58.8\% WER, against 21.7\% and 31.9\% for checkpoint initialization (primary paired comparison, $+23.2$ percentage points CER, 95\% CI $[+13.0, +34.9]$, $p=1.0\times10^{-7}$). A sensitivity analysis restricted to the 22 folds that converged gave the same direction with a smaller effect: 32.4\% CER and 49.4\% WER versus 22.6\% and 33.0\% ($+9.7$ percentage points CER, 95\% CI $[+6.5, +13.2]$, $p=3.3\times10^{-6}$), with random initialization better in 1 of 22 folds. The released checkpoint therefore contributed both optimization stability and decoding accuracy that multi-subject pretraining alone did not recover at this cohort size.

\subsection{Three Minutes of Calibration Yielded Similar Error Rates to Full Calibration}
\label{sec:results_sparse_calibration}

\begin{figure*}[!t]
\centering
\includegraphics[width=\textwidth]{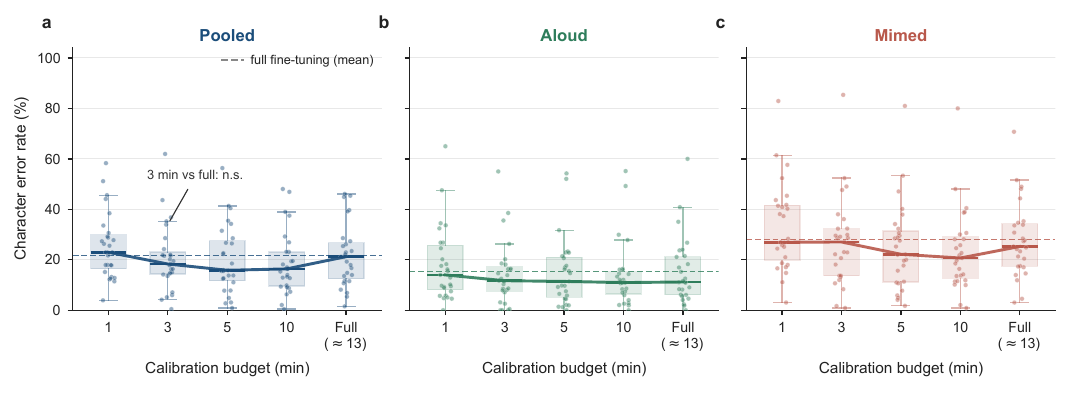}
\caption{Sparse target-subject calibration across 27 held-out subjects for Pooled, Aloud, and Mimed trials. Boxes show the median and interquartile range, dots show subjects, and lines connect budget medians; the dashed line marks the full-fine-tuning mean. Statistical comparisons are reported in Table~\ref{tab:sparse_calibration}.}
\label{fig:sparse_calibration_full}
\end{figure*}

We next asked how much held-out-subject calibration data is required. We repeated the Pretrain + fine-tune (MLP) experiment with different subsets of the target-subject calibration data. The sweep showed a steep initial improvement followed by a broad plateau (Fig.~\ref{fig:sparse_calibration_full}; Table~\ref{tab:sparse_calibration}). One minute of calibration produced most of the improvement but remained significantly worse than full calibration (25.4\% CER, 36.6\% WER; $p=0.003$ and $p=0.009$). With 3 min, the model reached 20.5\% CER and 31.7\% WER, compared with 21.7\% CER and 31.9\% WER using the full approximately 13-min pool; neither difference was statistically significant ($p=0.27$ and $p=0.88$). Five minutes was likewise not significantly different from full calibration ($p=0.26$ and $p=0.11$). Ten minutes yielded the lowest mean CER (18.6\%), but we do not interpret this as evidence that using less calibration data is intrinsically better, because the experiment fixed the number of epochs rather than the number of optimizer updates. No statistically significant difference was detected between 3 min and full calibration.

The sweep fixed the number of training epochs rather than optimizer updates. Because the complete calibration pool contained more trials per epoch, calibration duration and update count varied together. We therefore interpret the experiment as identifying 3 min as the shortest tested budget whose error rates were not significantly different from full calibration. The experiment does not establish equivalence among the 3-, 5-, 10-min, and full budgets, which were not compared with one another, nor does it show whether additional data would be beneficial under a matched-update schedule.

\begin{table*}[!t]
\renewcommand{\arraystretch}{1.15}
\caption{Sparse calibration sweep with the Pretrain + fine-tune (MLP) model ($n=27$)}
\label{tab:sparse_calibration}
\centering
\footnotesize
\resizebox{\textwidth}{!}{%
\begin{tabular}{lccccccc}
\toprule
Calibration budget & $n$ & CER (\%) & CER Std & WER (\%) & WER Std & $p_{\mathrm{CER}}$ vs.\ full & $p_{\mathrm{WER}}$ vs.\ full \\
\midrule
1 min & 27 & 25.4 & 13.1 & 36.6 & 16.3 & 0.003 & 0.009 \\
3 min & 27 & 20.5 & 13.2 & 31.7 & 17.5 & 0.27 & 0.88 \\
5 min & 27 & 19.9 & 13.5 & \textbf{28.4} & 17.2 & 0.26 & 0.11 \\
10 min & 27 & \textbf{18.6} & 12.5 & 28.6 & 16.9 & 0.030 & 0.049 \\
Full fine-tuning ($\approx$13 min) & 27 & 21.7 & 12.5 & 31.9 & 17.0 & -- & -- \\
\bottomrule
\end{tabular}}
\begin{flushleft}
\footnotesize
Values are macro means; Std is the subject-level sample standard deviation. The last two columns report descriptive, unadjusted paired two-sided Wilcoxon tests of each budget against full fine-tuning; no multiple-comparison correction is applied across budgets. All budgets used equal epoch counts, not equal optimizer-update counts.
\end{flushleft}
\end{table*}

Mimed CER remained higher throughout the sweep; at 5 min it was 24.8\% versus 15.0\% for Aloud.

\subsection{No Detectable Benefit of the Subject-Specific MLP}
\label{sec:results_adapter}

We next asked whether the subject-specific MLP contributes to the personalization gain. Adding the MLP changed pooled CER by $-0.8$ percentage points (95\% confidence interval $[-2.9, +1.2]$; $p=0.44$), with the MLP better for 15 of the 27 subjects, no MLP better for 11, and 1 exact tie. The same conclusion held for Mimed trials (27.9\% vs.\ 29.3\% CER for MLP vs.\ no MLP; MLP lower in 14/27 subjects, $p=0.68$), as well as for Aloud trials. The confidence interval spans zero and admits effects of at most a couple of points in either direction, so we find no detectable benefit from the adapter.

The personalization gain comes from multi-subject pretraining followed by supervised target-subject fine-tuning, not from the adapter architecture: standard full-model fine-tuning of the shared decoding model is sufficient. We tested this only with the 26-subject pretraining pool, so we cannot say whether the adapter would earn its place when fewer subjects are available for pretraining.

\subsection{Performance Improves with the Number of Pretraining Subjects}
\label{sec:results_scaling}

\begin{figure}[!ht]
\centering
\includegraphics[width=\columnwidth]{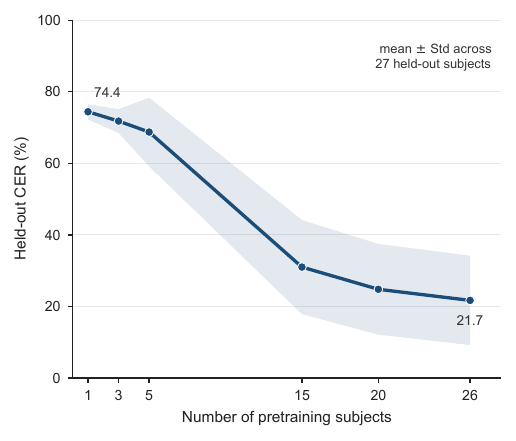}
\caption{Held-out CER against the number of pretraining subjects for the Pretrain + fine-tune (MLP) model, pooling Aloud and Mimed trials. All 27 held-out subjects are evaluated at every pool size using randomly selected non-held-out subjects. The line shows the macro mean and the shaded band shows $\pm$1 subject-level Std. The 26-subject endpoint corresponds to the Pretrain + fine-tune (MLP) condition in Table~\ref{tab:full_cohort_benchmark}.}
\label{fig:pretraining_subject_scaling}
\end{figure}

The pretraining subject-count scaling experiment (Fig.~\ref{fig:pretraining_subject_scaling}) directly tested whether increasing the number of pretraining subjects improves performance on a held-out subject even when the pretrained model is fine-tuned with target-subject calibration data. We used random pools of 1, 3, 5, 15, 20, and 26 non-held-out subjects. Every held-out subject was tested at every pool size, and the pools were drawn at random rather than ranked by the phoneme-based sEMG prediction score.
The 26-subject endpoint is the Pretrain + fine-tune (MLP) condition of Table~\ref{tab:full_cohort_benchmark}.

Macro mean CER fell from 74.4\% with one pretraining subject to 21.7\% with 26. The endpoint difference was 52.7 percentage points (95\% CI $[48.3, 57.1]$, $p=1.5\times10^{-8}$), and all 27 held-out subjects improved between these endpoints. Within each subject, the rank correlation between pool size and CER was negative for all 27 (median Spearman $\rho=-1.00$). The decrease was not step-by-step monotonic for every subject: 16 of 27 improved at every increase in pool size, while the remaining 11 showed excursions at the small pool sizes, where the model was barely usable and its errors were erratic. The prespecified comparisons $1\rightarrow5$, $5\rightarrow15$, $15\rightarrow20$, and $20\rightarrow26$ were each significant, so the curve had not flattened by 26 subjects. The steepest part lay between five and 15 pretraining subjects, where CER dropped by 37.7 percentage points. Below five subjects, the model was close to unusable (CER above 68\%, WER at or near 100\%), and target-subject calibration did not offset the weak pretraining result. This scaling trend supports the main conclusion that pretraining on many subjects drives cross-subject personalization.

\subsection{Performance on Unseen Sentences Remains Limited}
\label{sec:results_unseen_sentences}

We next tested unseen sentences across all 27 subjects using the Pretrain + fine-tune (MLP) model. This test reused the same 269 held-out trials as the closed-corpus evaluation, comprising the reserved Aloud and Mimed trials of the five evaluation sentences for each subject. In both analyses, those sentences were excluded from target-subject calibration. For the unseen-sentence test, they were additionally excluded from multi-subject pretraining. The two rows of Table~\ref{tab:unseen_sentences} therefore used identical held-out trials but differed in pretraining exposure and training budget. The larger epoch budget for the unseen-sentence models did not prevent the observed degradation.

\begin{figure}[!ht]
\centering
\includegraphics[width=\columnwidth]{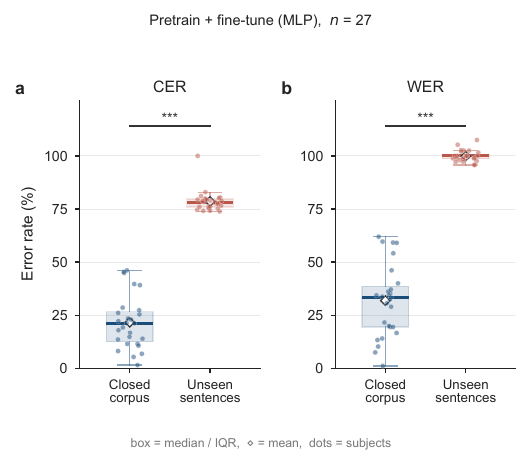}
\caption{Closed-corpus versus unseen-sentence performance for the same 27 held-out subjects, using the Pretrain + fine-tune (MLP) model with Aloud and Mimed trials pooled. Panels show (a) CER and (b) WER. Boxes indicate the median and interquartile range, diamonds indicate macro means, and dots indicate individual subjects. Stars denote paired two-sided Wilcoxon signed-rank comparisons ($^{***}p<0.001$).}
\label{fig:unseen_sentences}
\end{figure}

Performance deteriorated sharply on the identical held-out trials, and every one of the 27 subjects performed worse (Table~\ref{tab:unseen_sentences}; Fig.~\ref{fig:unseen_sentences}). Thus, successful adaptation to a new subject in the closed-corpus evaluation should not be interpreted as open-content decoding.

\begin{table}[!ht]
\renewcommand{\arraystretch}{1.15}
\centering
\caption{Unseen-sentence stress test ($n=27$)}
\label{tab:unseen_sentences}
\resizebox{\columnwidth}{!}{%
\begin{tabular}{lccccc}
\toprule
Evaluation & $n$ & CER (\%) & CER Std & WER (\%) & WER Std \\
\midrule
Held-out trials, closed-corpus condition & 27 & \textbf{21.7} & 12.5 & \textbf{31.9} & 17.0 \\
Unseen sentences & 27 & 78.6 & 4.9 & 99.9 & 2.6 \\
\bottomrule
\end{tabular}}
\begin{flushleft}
\footnotesize
Both rows use the Pretrain + fine-tune (MLP) model on the identical 269 held-out test trials and differ only in whether the evaluation sentences appeared during multi-subject pretraining and in training budget (400 versus 800 epochs). Std is the subject-level sample standard deviation across the 27 held-out subjects. WER can exceed 100\% because it is an edit distance normalized by reference length, so insertions are unbounded.
\end{flushleft}
\end{table}

\subsection{Closed-Set Template Baselines}
\label{sec:results_closed_set}

Because the evaluation uses a closed 50-sentence corpus, we asked how much of the decoding performance is explained by a sentence prior over this corpus rather than by genuine sub-sentence decoding. On the same held-out trials, which were never used in training, we evaluated two model-independent closed-set baselines: a uniform chance baseline, whose phrase-selection accuracy over the 50-sentence corpus is 2\%, and a nearest-template recognizer built from each held-out subject's own training trials. Under the closed-corpus protocol these template trials include the non-reserved repetitions of the evaluation sentences, so all 50 corpus sentences are available as candidate labels, while the reserved test trials themselves are excluded from the template set. Test and template trials were processed identically: each channel was full-wave rectified, smoothed with a moving average spanning approximately 5\% of the trial duration, linearly resampled to a fixed 50 frames, and z-scored, and the eight channels were then concatenated into a single feature vector. Each test trial was assigned the sentence of its nearest template under Euclidean distance (1-nearest-neighbour). The nearest-template recognizer performed well above chance, confirming that the corpus is partially separable at the sentence level from raw EMG activity alone. Across all 27 subjects, the personalized decoder reduced CER by 38.6 percentage points, or 64.0\% relative to the nearest-template baseline (Table~\ref{tab:closed_set}). This result shows that the learned decoder substantially outperformed a simple within-subject nearest-template recognizer. However, it does not by itself establish compositional or open-content decoding.

\begin{table}[!ht]
\renewcommand{\arraystretch}{1.15}
\centering
\caption{Closed-set lower-bound baselines ($n=27$)}
\label{tab:closed_set}
\resizebox{\columnwidth}{!}{%
\begin{tabular}{lccc}
\toprule
Method & $n$ & CER (\%) & WER (\%) \\
\midrule
Chance (uniform over 50 sentences) & 27 & 89.6 & 115.1 \\
Closed-set nearest-template (EMG envelope 1-NN) & 27 & 60.3 & 76.1 \\
Personalized decoder (Pretrain + fine-tune (MLP)) & 27 & \textbf{21.7} & \textbf{31.9} \\
\bottomrule
\end{tabular}}
\begin{flushleft}
\footnotesize
Values are macro-averaged across all 27 held-out subjects on identical closed-corpus trials. The nearest-template recognizer selects among all 50 corpus sentences (top-1 phrase-selection accuracy 32.6\%; chance 2\%). WER exceeds 100\% for the chance row because insertion errors are unbounded.
\end{flushleft}
\end{table}

\subsection{Exploratory Encoding--Decoding Score Association}
\label{sec:results_encdec_corr}

This exploratory analysis asks whether a subject's phoneme-based sEMG encoding score, computed independently of decoding, predicts how well that subject can be decoded. Here \emph{encoding} denotes a forward model that predicts sEMG activity from time-aligned speech features, whereas decoding predicts speech content from sEMG. We reused subject-level scores from our prior study \cite{le2026semg_encoding}, in which an elastic-net multivariate temporal-response model \cite{crosse2016mtrf,zou2005elasticnet} predicted each Aloud sEMG envelope from phoneme one-hot features; performance was the Pearson correlation between predicted and recorded envelopes on held-out sentences, averaged across the eight channels. The encoding models were not refit here. The same score also defines a curated top-10 subset, namely the ten subjects with the highest scores (P5, P6, P7, P8, P10, P17, P21, P24, P28, and LP2). By construction this subset had a higher median encoding score than the remaining scored subjects (0.470 vs.\ 0.388); total EMG durations did not differ significantly between the two groups (Mann--Whitney $p=0.598$). Ranking instead by an articulatory encoding score selects the identical ten subjects within this cohort, so the subset does not depend on which encoding score is used.

We related this score to decoding accuracy, quantified as $1-\mathrm{CER}$, on the LOSO pretrained-plus-fine-tuned decoder. Among the 22 held-out subjects with valid scores, the association was weakly positive but not statistically significant: two-sided Pearson $r=0.23$ ($p=0.30$; 95\% Fisher-$z$ confidence interval $[-0.21,0.60]$) and two-sided Spearman $\rho=0.12$ ($p=0.59$). On the top-10 subset the point estimate was near zero (Pearson $r=0.01$, $p=0.97$; Spearman $\rho=-0.03$, $p=0.93$), but with $n=10$ the 95\% confidence interval spans $[-0.62,0.64]$, so this analysis is underpowered and cannot establish the absence of an association. We therefore report that the prior encoding score did not measurably explain subject-level decoding difficulty in this corpus, without treating the top-10 result as evidence of no relationship.

\section{Discussion}

\subsection{Scalable Pretraining and Short Fine-Tuning Drive Personalization}

The main finding is that closed-corpus cross-subject sEMG decoding benefits from multi-subject pretraining followed by supervised target-subject fine-tuning. In the 27-subject LOSO analysis, fine-tuning on the held-out subject substantially outperformed transfer without calibration. Two results support the role of multi-subject pretraining: direct checkpoint fine-tuning remained poor despite using the broader target-subject calibration pool, and performance improved as the pretraining pool grew. Because the pretraining pool was randomly selected from the remaining non-held-out subjects for each fold and pool size, the trend does not depend on one fixed ordering of subjects. The Checkpoint + fine-tune condition is a protocol control showing that, under the same optimization schedule, checkpoint initialization followed by target-subject-only fine-tuning was insufficient to reproduce the benefit of multi-subject pretraining. It is not intended as a comparison against a separately optimized single-subject system. The converse control, multi-subject pretraining from random initialization, shows that the dependence runs in both directions: neither ingredient was sufficient alone at this data scale, since removing the released initialization cost 9.7 percentage points of CER on the folds that converged and prevented convergence altogether in 5 of 27 folds. Personalization here therefore rests on the combination of a pretrained initialization, multi-subject pretraining, and target-subject calibration.

The calibration sweep provides a practical estimate for a future setup session. In the present cohort of healthy, speech-typical participants recorded with a standardized eight-channel montage, 3 min of target-subject calibration yielded no statistically significant difference from the full approximately 13-min calibration pool. Ten minutes produced the lowest mean CER, but the fixed-epoch design does not support the conclusion that using less data is intrinsically better, because calibration duration and optimizer-update count varied together. The 3-min condition therefore provides a practical short-calibration operating point under the present protocol rather than a universal minimum calibration requirement, and motivates testing a short setup session in future systems built around a constrained set of phrases. The calibration-duration estimate is specific to the present protocol and may change for users with speech impairment, for different electrode montages, after electrode replacement, or across recording sessions; longitudinal and electrode-shift robustness were not tested here.

\subsection{No Detectable Benefit from the Subject-Specific Adapter}

The 27-subject comparison does not show a benefit from the subject-specific MLP. Fine-tuning with and without the adapter gave similar error rates, and the confidence interval for their difference spans zero. The subject-count experiment used the MLP model at every pool size, so it cannot show whether the adapter behaves differently with smaller pretraining pools. The conclusion supported by both analyses is that pretraining on many subjects, followed by supervised fine-tuning on the new subject, improves closed-corpus decoding.

\subsection{Scope of the Closed-Corpus Results}

The unseen-sentence experiment defines where the main personalization result applies. It used the same test trials and target-subject calibration split, but additionally removed the trials corresponding to the held-out subject's evaluation sentences from multi-subject pretraining. Every held-out subject performed worse. We use this experiment only to define the scope of the closed-corpus analysis. The main result applies within this corpus; it does not establish open-content decoding.


\subsection{Relation to Prior Cross-Subject sEMG Work}

Recent cross-user sEMG work on non-speech tasks has achieved transfer with little or no calibration \cite{kaifosh2025generic}. In sentence-level speech decoding, our results show that short calibration remains necessary. Prior sEMG studies have used domain alignment or meta-learning to reduce differences between users \cite{wand2018domain,prorokovic2019meta,zhang2023crosssubject,cui2024adversarial}. We instead use multi-subject pretraining followed by supervised fine-tuning, a paradigm already shown to transfer across speakers for EMG-to-speech conversion \cite{scheck2024crossspeaker}. Closest to the present work, Zeng et al.~\cite{zeng2025crosssubject} calibrated a cross-subject sEMG-to-speech system with varying amounts of target-subject data in an 11-subject cohort. Our study differs in the decoding target, character sequences scored by CER and WER rather than synthesized speech, in cohort size and per-subject data budget, and in that we vary the size of the pretraining cohort as well as the calibration duration. Gaddy and Klein~\cite{gaddy2020digitalvoicing} established single-subject open-vocabulary silent-speech decoding; here their architecture is evaluated in a cross-subject, closed-corpus setting. Because these studies use different corpora, electrode montages, and tasks, we compare them qualitatively rather than by their reported error rates.

\subsection{Implications for Rehabilitation and Neural Engineering}

Short calibration may be useful in future constrained sEMG communication systems, complementing prior work on practical silent-speech system design \cite{deng2014practical}. The unseen-sentence result limits this interpretation to a fixed sentence inventory. Longitudinal, electrode-placement, clinical-population, and open-content tests are needed before deployment.

\subsection{Limitations}

This study has several limitations. First, the primary analysis uses a closed 50-sentence corpus in which sentences recur during training. Performance on unseen sentences remained poor, so the main result should not be interpreted as open-vocabulary speech decoding. Second, the analyses are offline and do not measure communication rate, processing latency, user feedback, or closed-loop adaptation. Third, the participants are healthy adults rather than users with speech impairments.

The main models were initialized from a released single-subject checkpoint trained with a different eight-channel montage. The matched random-initialization condition reported in Section~\ref{sec:results_full_cohort} shows that this dependence is substantial rather than incidental: at the present cohort size, multi-subject pretraining alone neither matched the accuracy of checkpoint initialization nor reliably converged. We did not determine whether this dependence would persist with a larger pretraining cohort, nor did we search for an optimization schedule that stabilizes training from random initialization; the schedule was held fixed across conditions by design, so the comparison shows what the released checkpoint contributes under this schedule rather than an upper bound on what random initialization could achieve if tuned separately. Finally, longitudinal variability, electrode replacement, and electrode-shift robustness were not tested \cite{ameri2020electrode_shift}.

\subsection{Future Work}

Future work should evaluate longitudinal and electrode-shift robustness, collect larger and more varied sentence sets, and test training objectives that improve phoneme- or articulatory-level generalization rather than sentence memorization. Clinical studies should also determine whether the benefit of multi-subject pretraining and the short calibration requirement observed in healthy participants transfer to assistive-communication users.

\section{Conclusion}

Our results show that multi-subject pretraining supports closed-corpus sEMG decoder personalization when only limited data are available from each participant. Across 27 held-out participants, target-subject fine-tuning after multi-subject pretraining substantially outperformed zero-shot transfer and direct target-subject fine-tuning of the released checkpoint under the same training schedule, confirming that target-subject calibration remains necessary. The matched random-initialization control was substantially less accurate and failed to converge under the fixed schedule in 5 of 27 folds, indicating that the released checkpoint contributed both optimization stability and decoding accuracy in this limited-data regime. Under the fixed-epoch schedule, 3 min of calibration was not significantly different from the full approximately 13-min calibration pool, while a subject-specific MLP adapter provided no detectable advantage. These findings are specific to a cohort of healthy, speech-typical participants recorded with a standardized eight-channel montage and to a fixed 50-sentence corpus: decoding deteriorated sharply when the evaluation sentences were withheld from all sEMG model-training data, so reliable zero-shot and open-content decoding remain unresolved.

\section*{Ethics Statement}

The experimental procedures involving human subjects described in this study were approved by New York University's Institutional Review Board under the protocol ``Auditory Sequences: Psychophysics and Neurophysiology'' (IRB-FY2020-4157).

\section*{Data and Code Availability}

The de-identified sEMG dataset, associated annotations, and code used in this study will be made publicly available upon publication at \url{https://huggingface.co/datasets/whisperle/multisubject-semg-text-v1}. Raw audio recordings and other potentially identifiable participant data will not be publicly released due to institutional and participant-privacy restrictions.

\section*{Acknowledgments}

This work was supported in part by the NYU Discovery Research Fund for Human Health and in part by the U.S. National Science Foundation under Award 2309057. We thank Dr. Pablo Ripolles for his support related to the institutional review protocol under which the data were collected. We acknowledge the NYU High Performance Computing (HPC) resources, services, and staff for supporting the computational work reported in this study. The authors used OpenAI ChatGPT and Anthropic Claude Code for language editing, code-assisted manuscript revision, and consistency checking of reported values against the underlying result files. All AI-assisted text, code, analyses, and reported values were independently verified and edited by the authors, who take full responsibility for the manuscript content.
\bibliographystyle{unsrt}
\bibliography{bibtex/bib/semg_loso}

\end{document}